\documentclass[runningheads]{llncs}
\usepackage{graphicx}
\usepackage{amsmath}
\usepackage{amssymb}
\usepackage{algorithm2e}
\usepackage{booktabs}
\RestyleAlgo{ruled}

\usepackage{subcaption}
\usepackage{orcidlink}
\usepackage{cleveref}
\Crefformat{figure}{#2Fig.~#1#3}

\usepackage[utf8]{inputenc}
\usepackage{pgfplots}
\DeclareUnicodeCharacter{2212}{−}
\usepgfplotslibrary{groupplots,dateplot}
\usetikzlibrary{patterns,shapes.arrows}
\pgfplotsset{compat=newest}
\pgfplotsset{compat=newest}
\pgfplotsset{legend image post style={black},legend style = {text=black}}
\usepackage{hyperref}

\begin{document}
\title{Fusing Forces: Deep-Human-Guided Refinement of Segmentation Masks\thanks{This project has received funding from the Austrian Science Fund/Österreichischer Wissenschaftsfonds (FWF) under grant agreement No. P~33721.}}
\titlerunning{Deep-Human-Guided Refinement of Segmentation Masks}
%
\author{Rafael Sterzinger\orcidlink{0009-0001-0029-8463} \and
Christian Stippel\orcidlink{0000-0003-0482-902X} \and
Robert Sablatnig\orcidlink{0000-0003-4195-1593}}
\authorrunning{R. Sterzinger et al.}
%
\institute{Computer Vision Lab, TU Wien, Vienna, AUT
\email{\{firstname.lastname\}@tuwien.ac.at}}
\maketitle              
\begin{abstract}
Etruscan mirrors constitute a significant category in Etruscan art, characterized by elaborate figurative illustrations featured on their backside.
A laborious and costly aspect of their analysis and documentation is the task of manually tracing these illustrations.
In previous work, a methodology has been proposed to automate this process, involving photometric-stereo scanning in combination with deep neural networks.
While achieving quantitative performance akin to an expert annotator, some results still lack qualitative precision and, thus, require annotators for inspection and potential correction, maintaining resource intensity.
In response, we propose a deep neural network trained to interactively refine existing annotations based on human guidance.
Our human-in-the-loop approach streamlines annotation, achieving equal quality with up to 75\% less manual input required.
Moreover, during the refinement process, the relative improvement of our methodology over pure manual labeling reaches peak values of up to 26\%, attaining drastically better quality quicker. 
By being tailored to the complex task of segmenting intricate lines, specifically distinguishing it from previous methods, our approach offers drastic improvements in efficacy, transferable to a broad spectrum of applications beyond Etruscan mirrors.
\keywords{Binarization \and Interactive Segmentation \and Human-in-the-Loop \and Etruscan Art \and Cultural Heritage}
\end{abstract}

\begin{figure}[t]
    \begin{subfigure}{0.31\textwidth}
        \centering
		\includegraphics[width=.9\textwidth]{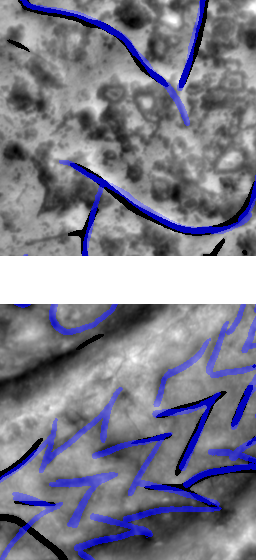}
        \caption{Initial Prediction $\mathbf{Y}$}
	\end{subfigure}
    \hfill
 	\begin{subfigure}{0.31\textwidth}
        \centering
		\includegraphics[width=.9\textwidth]{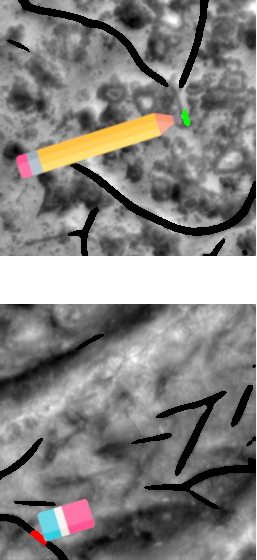}
        \caption{Human Interaction $\mathbf{\Delta}$}
	\end{subfigure}
    \hfill
 	\begin{subfigure}{0.31\textwidth}
        \centering
		\includegraphics[width=.9\textwidth]{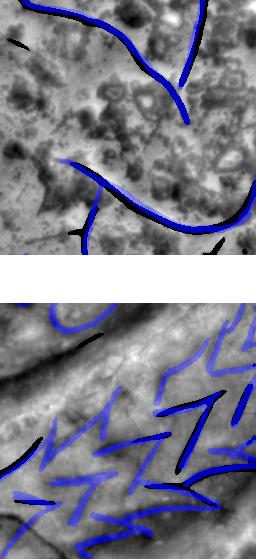}
        \caption{Refined Prediction $\mathbf{Y}'$}
	\end{subfigure}
 \caption{Illustrating interactive refinement of segmentation masks: Starting from an initial segmentation $\mathbf{Y}$, the user can add ($\mathbf{\Delta}^+$) or erase ($\mathbf{\Delta}^-$) parts to bring it closer to the ground truth $\mathbf{Y}^*$ (in blue), creating an updated mask $\mathbf{Y}^\mathbf{\Delta}$.
 Next, using a separate model conditioned on the human input $\mathbf{\Delta}$ and $\mathbf{Y}$, we aim that for the refined segmentation $\mathbf{Y}'$ it holds that~$||\mathbf{Y}'-\mathbf{Y}^*||_1<||\mathbf{Y}^\mathbf{\Delta}-\mathbf{Y}^*||_1$.}
\label{fig:obtions}
\end{figure}

\section{Introduction}
\begin{figure}[h]
	\centering
    \begin{subfigure}{0.48\textwidth}
	        \includegraphics[width=\textwidth]{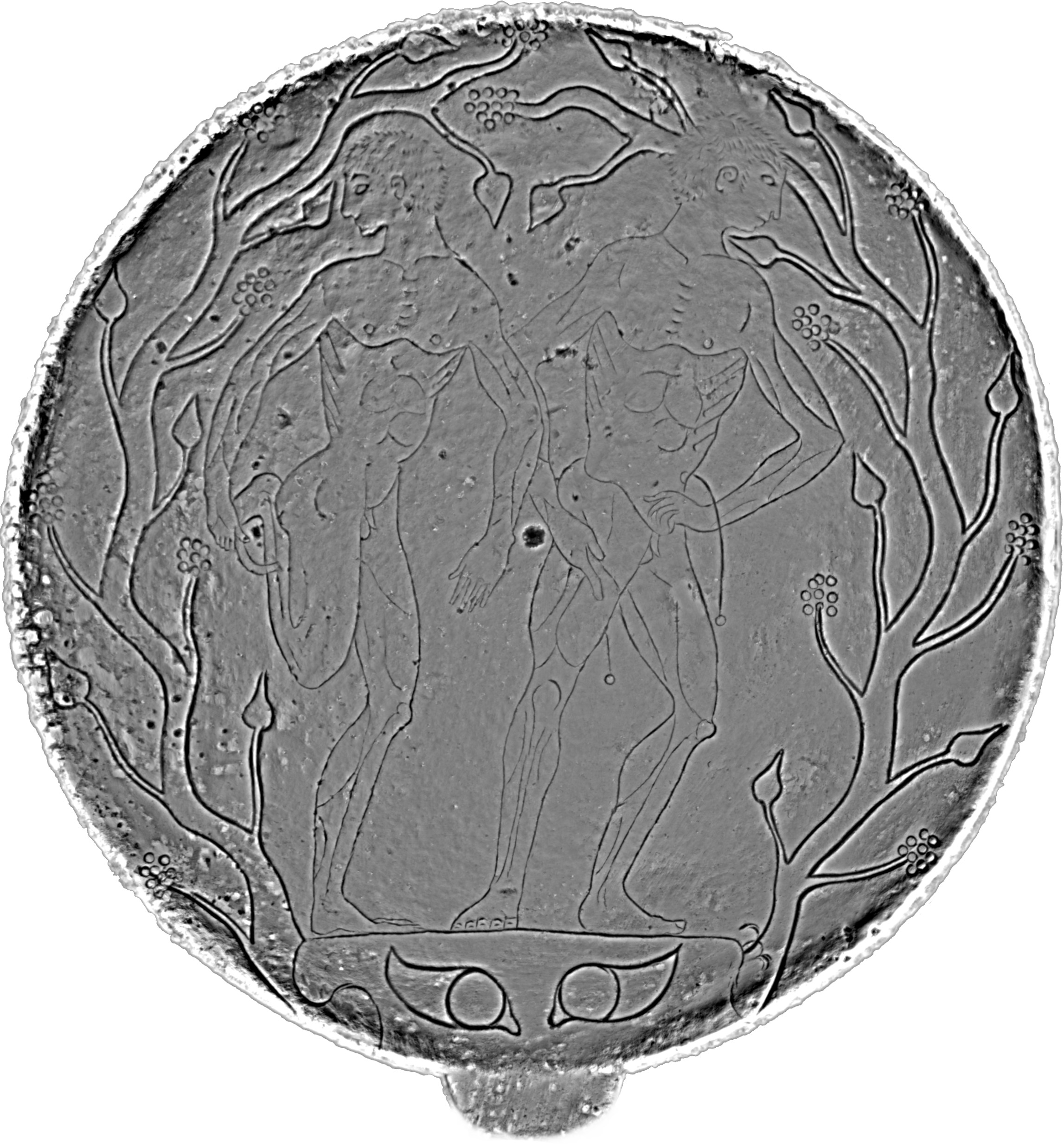}
\caption{High-Pass Filtered Depth Map}
    \end{subfigure}
    \hfill
	\begin{subfigure}{0.48\textwidth}
	        \includegraphics[width=\textwidth]{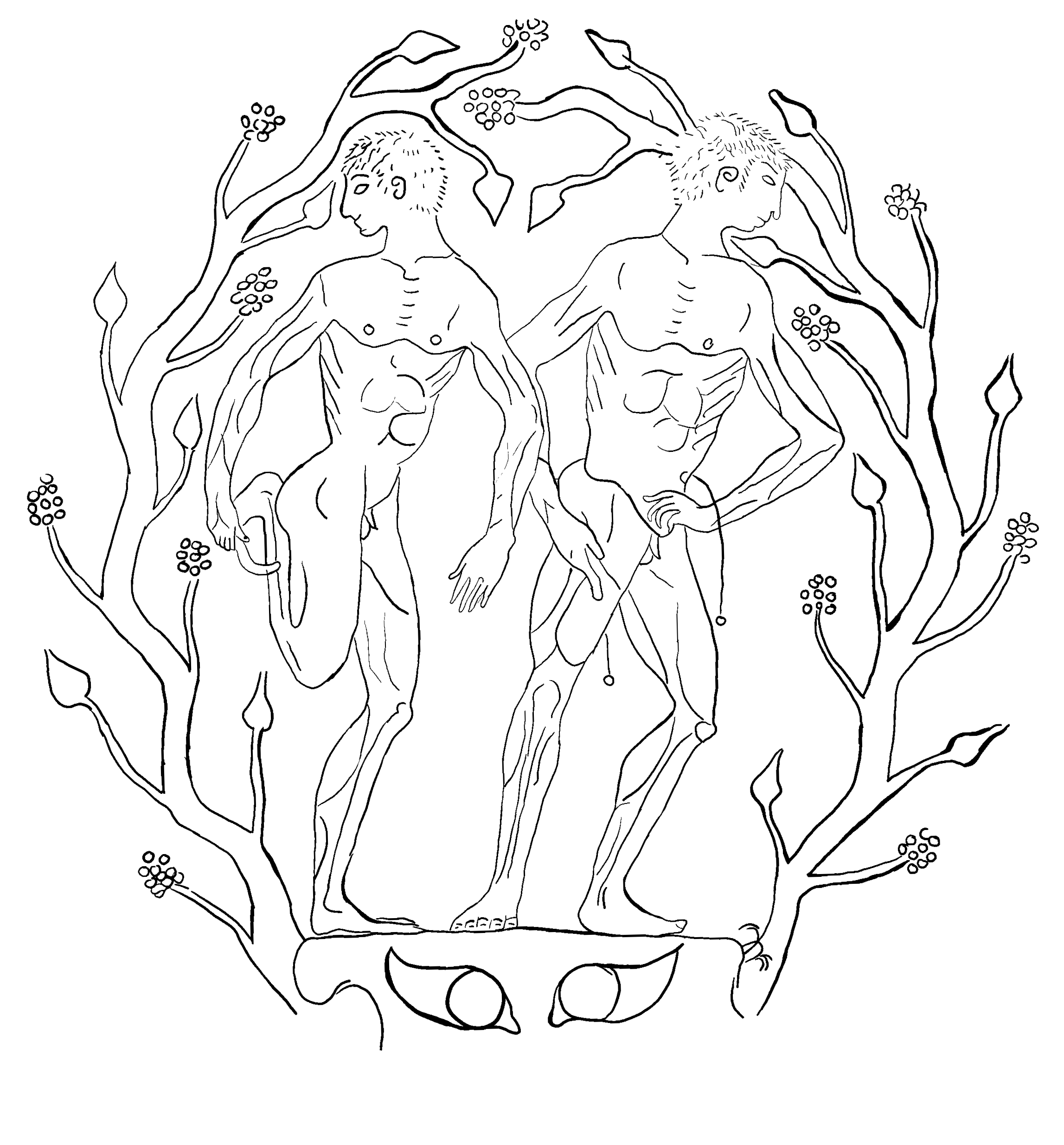}
	        \caption{Ground Truth Segmentation Mask}
    \end{subfigure}
    \caption{Etruscan mirrors typically feature scenes from Greek mythology.
    During their examination, archaeologists seek to extract the drawings for visualization.}
    \label{fig:input}
\end{figure}

With more than 3,000 identified specimens, Etruscan hand mirrors represent one of the biggest categories within Etruscan art.
On the front, these ancient artworks feature a highly polished surface, whereas, on the back, they typically depict engraved and/or chased figurative illustrations of Greek mythology~\cite{sindy_2023}.
A primary component of their examination involves the labor- and cost-intensive task of manually tracing the artworks; an exemplary mirror is illustrated in \Cref{fig:input} together with the sought-after tracing. 

In previous works, Sterzinger et al.~\cite{sterzinger_drawing_2024} propose a methodology to automate the segmentation process through photometric-stereo scanning in combination with deep learning; expediting the process of manual tracing and contributing to increased objectivity.
Although their segmentation model --~trained on depth maps of Etruscan mirrors to recognize intentional lines over scratches~-- already quantitatively achieves performance on par with an expert annotator, in some instances it lags behind.
Based on this, manual inspection and potential refinement by humans are still required, therefore, although alleviated, the tracing remains resource-intensive.

In this paper, we continue their line of work and propose a methodology to simplify the remaining required refinement by adding interactivity to the process: Starting from an initial prediction, we aim to reach qualitatively satisfying results as quickly as possible while keeping necessary labor to a minimum.
We achieve this by training a deep neural network to refine the initial segmentation based on a series of hints, i.e., parts being added or erased, illustrated in \Cref{fig:obtions}.

In summary, our contribution entails the development of an interactive refinement network for improved annotation results obtained in less time, requiring less labor.
Compared to refining the initial segmentation manually, fusing forces and performing the refinement interactively offers not only a drastic reduction in labor (up to -75\%) but also expedites the process by attaining significant relative performance improvements over manual labeling (up to +26\%).
We differentiate ourselves from prior work by proposing a methodology tailored specifically to the task of segmenting intricate lines scattered across, in our case, Etruscan mirrors versus, e.g., segmenting locally-concentrated hepatic lesions~\cite{amrehn_ui-net_2017}.
Additionally, instead of starting from scratch, we start from an initial prediction, a step required due to the non-locality of lines as otherwise labor would be drastically higher.

Finally, we provide public access to both the code and data utilized in this work (see \href{https://github.com/RafaelSterzinger/etmira-interaction}{github.com/RafaelSterzinger/etmira-interaction}) to promote transparency and reproducibility.

\section{Related Work}

\paragraph{Segmentation:} In the field of image segmentation, techniques are led by advanced deep learning architectures such as the UNet~\cite{ronneberger_u-net_2015}, DeepLabV3++~\cite{chen_encoder-decoder_2018}, Pyramid Attention Network~\cite{li_pyramid_2018}, etc. 
These advancements are particularly propelled by industries where precise segmentation is paramount: For example, in medical imaging, intricate segmentations are crucial for identifying vascular structures within the retina, a crucial aspect for diagnosing retinal diseases~\cite{li_iternet_2020}.

\paragraph{Photometric Stereo:}
When considering historical artifacts where the content of interest is engraved or chased into the object, as is the case with Etruscan mirrors, instead of RGB, modalities that capture surface details are potentially better suited.
Photometric Stereo (PS), a technique introduced by Woodham~\cite{woodham1980photometric}, allows for capturing such details, providing insights into the surface geometry of an object.
For instance, McGunnigle and Chantler~\cite{mcgunnigle2003resolving} extract handwriting on paper based on depth profiles.
In addition to this, PS is also employed, e.g., to detect cracks in steel surfaces~\cite{LandstromThurleyJonsson2013}, extract leaf venation~\cite{ZhangHansenSmithSmithGrieve2018}, or detect air voids in concrete~\cite{TaoGongWangLuoQiuHuang2022}.
In the context of Etruscan mirrors, Sterzinger et al. \cite{sterzinger_drawing_2024} resort to a deep-learning-based segmentation approach due to the damage these mirrors have sustained.
By integrating PS-scanning with deep segmentation, they learn to recognize intentional lines over scratches.

\paragraph{Interactive Segmentation:} Independent of the segmentation methodology employed, resulting masks might not meet performance requirements and, therefore, require correction.
With this regard, Li et al.~\cite{li_iternet_2020}, introduce IterNet, a UNet-based iterative approach to enforce connectivity of retinal vessels post segmentation, requiring no-external input.
Similarly, interactive methods exist that incorporate human expertise within the process.
Xu et al.~\cite{xu_deep_2016} and Mahadevan et al.~\cite{mahadevan_iteratively_2018} focus on object segmentation based on mouse clicks.
On the other hand and closest to our work, Amrehn et al.~\cite{amrehn_ui-net_2017}, propose an approach that refines the segmentation based on pictorial scribbles for hepatic lesions.

\section{Methodology}

In the following we will detail our methodology comprised of:
\begin{itemize}
    \item the dataset; general information, splitting the data into training, validation, and testing, as well as, the preprocessing of depth maps
    \item the simulation of human interaction; details on the statistics of engravings, acquiring individual line segments, and the procedure for adding and erasing 
    \item the architecture; describing the overall deep neural network used for refining the initial segmentation
\end{itemize}

\subsection{Dataset}

Our dataset includes a diverse array of Etruscan mirrors from public collections in Austria.
It consists of PS-scans of 59 mirrors, with 53 located at the Kunsthistorischen Museum (KHM) Wien and the remaining 6 scattered throughout Austria.
Annotations were acquired for 19 mirrors, encompassing 19 backsides and 10 fronts, resulting in a total of 29 annotated examples.
Notably, engravings predominantly adorn the backside to avoid interference with reflectance, however, they are also occasionally found on the front, albeit with less density, near the handle or around the border.
For information on the acquisition process, we refer the reader to Sterzinger et al.~\cite{sterzinger_drawing_2024}.

Dividing these annotations into training, validation, and test sets is challenging due to three factors: limited sample size, strong variations in the density of engravings, and overall mirror conditions.
Mirrors with dense engravings are prioritized for training due to the stronger learning signal they offer.
To ensure fair evaluation, we select three mirrors of different conditions and engraving densities for testing: one from Wels and two from the KHM Wien.
We create non-overlapping patches of size $512\times512$ pixels, shuffle, and split them in half to form the validation and test set of similar underlying distributions.
One outlier, characterized by a different art style (points instead of lines), is excluded, leaving 25 annotated samples for training. 

\subsubsection{Preprocessing}
With regards to preprocessing, we employ the depth modality (which worked best according to~\cite{sterzinger_drawing_2024}) and remove low frequencies.
We accomplish this by subtracting a Gaussian-filtered version of the depth map with values capped between $\mu \pm 3\sigma$.
In addition, employing the Segment Anything Model (SAM)~\cite{kirillov_segment_2023}, global segmentation masks are generated to identify the mirror object within a shot.
We use these masks to differentiate between mirror and non-mirror parts (see \Cref{fig:pipeline}; compare red versus green, top-left), for instance, to calculate per-channel means and standard deviations only on mirror parts which we use to normalize the input.

Addressing the lack of annotations, a per-patch inference approach is adopted.
For validation and testing, non-overlapping quadratic patches measuring $512\times512$ pixels are extracted.
Regarding our training data, we pad four pixels, since $6720 \equiv 0 \pmod{2240}$, to the original resolution ($8,964\times6,716$~pixels) to extract 25 overlapping tiles of size $2,988\times2,240$~pixels 
using a stride of half the size; tiles, containing no annotation, are discarded.
Diversifying the dataset for each epoch, ten patches per tile are extracted, all resized to dimensions of $256\times256$ pixels to streamline model complexity.

\subsection{Simulation of Human Interaction}

In order to simulate realistic human interactions, we first look into the statistics of the annotations included in the dataset; necessary to quantitatively capture human-stroke width.
Next, to refine initial predictions, we describe the process of filtering and correcting false positives and negatives.
Within this, we motivate and denote the algorithm used to extract line segments.
Tying all components together, we finally describe simulating human interaction: Starting from either false positives or negatives, we extract the largest error segment and provide a hint in the form of a line with width taken from the acquired statistics of the ground truth annotations.

\begin{algorithm}[h]
\SetAlgoLined
\KwData{ground truth $\mathbf{Y}^*$}
\BlankLine

\SetKwProg{myMethod}{def}{:}{}

\myMethod{get\_stroke\_widths$(\mathbf{Y}^*)$}{
distance\_map $\leftarrow$ euclidean\_distance\_transform$(\mathbf{Y}^*)$

gt\_skelet $\leftarrow$ skeletonize$(\mathbf{Y}^*)$

\Return distance\_map[gt\_skelet]
}

\caption{Calculating Stroke Widths for Statistics}
\label{alg:line_info}
\end{algorithm}

\subsubsection{Statistics}
With the goal of simulating realistic interaction, one crucial component to consider is the stroke width. 
For this, we look into the statistics of the annotations included in our dataset by extracting individual thickness, using \Cref{alg:line_info}: Starting from a binary mask, the ground truth $\mathbf{Y}^*$ in our case, we obtain distance information via the \texttt{euclidean\_distance\_transform} which, for each pixel, returns the Euclidean distance in pixels to the closest non-mask pixel.
Next, employing \texttt{skeletonize}~\cite{zhang1984fast}, we acquire a skeletonized version of the input (essential the center of lines), used to extract the thickness at each section.

\begin{figure}
    \centering
    \includegraphics[width=0.85\textwidth]{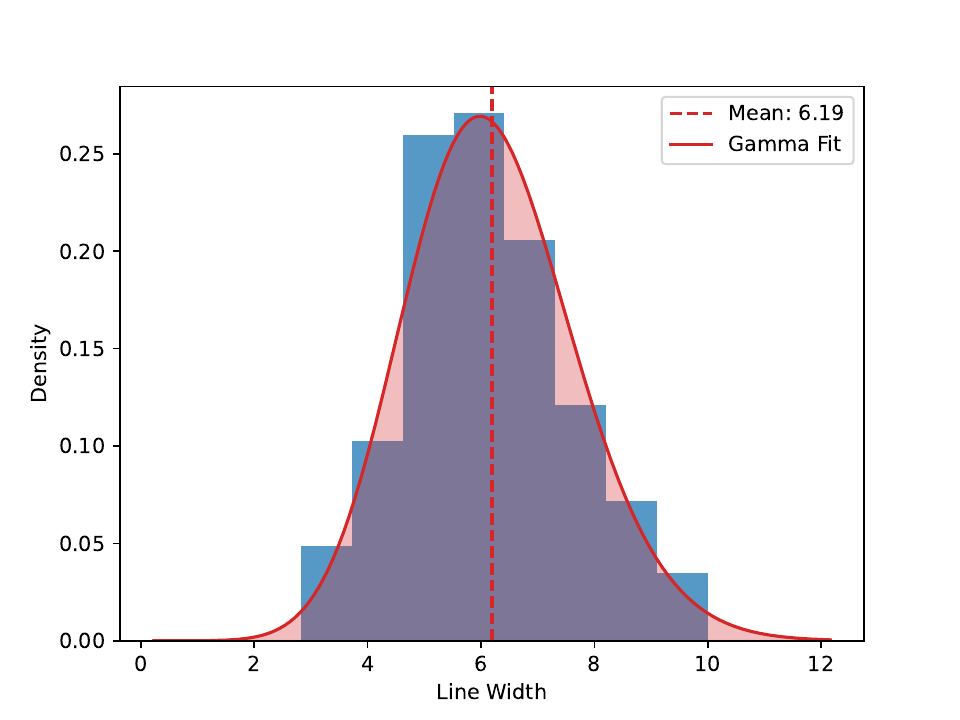}
    \caption{Illustration of the distribution of stroke widths: After removing outliers from our data, using the two-sigma rule, we fit a Gamma distribution (shape-parameter $a=49.13$, loc$=-4.28$, scale$=0.21$).
    }
    \label{fig:hist}
\end{figure}

From this information, we calculate initial $\mu$ and $\sigma$ of the collected line widths, which we use to remove outliers (long right tail) using the two-sigma rule, keeping values within two standard deviations, to obtain final $\mu=6.19$ and $\sigma=1.49$.
Based on this filtered set of stroke widths, we fit a Gamma distribution~$\mathcal{G}$ from which we can randomly sample realistic widths.
\Cref{fig:hist} visualizes the distribution of stroke widths as well as the fitted distribution $\mathcal{G}$.
 
\subsubsection{Operations}
In general, when an expert annotator is entrusted with the task of refining segmentation masks, one of two operations will be performed: adding missing parts or erasing superfluous ones.
Simulating these operations consists of multiple steps: (1)~finding an area that requires correction (we assume areas will be selected in decreasing order depending on the magnitude of correction required), (2)~deciding on an operation, and (3)~performing the operation.
In essence, however, steps (1) and (2) go hand in hand, i.e., when deciding on an area based on error, the operation to be performed is already clear.

Let $\mathbf{Y}=f_{init}(\mathbf{X}) \in \mathbb{B}^{H,W}$ be the initial segmentation mask produced by a baseline network~$f_{init}$ based on the depth map~$\mathbf{X} \in \mathbb{R}^{H,W}$.
Given that we generally work with patches, using this initial mask, we find the segment that requires the most correction w.r.t.\ our ground truth~$\mathbf{Y}^*$, considering the pFM~(formally introduced in \Cref{sec:evaluation}).


\begin{algorithm}
\SetAlgoLined
\KwData{ground truth $\mathbf{Y}^*$, prediction $\mathbf{Y}$, line statistics $\mu$ and $\sigma$}
\BlankLine

\SetKwComment{Comment}{// }{}
\SetKwProg{myMethod}{def}{:}{}
\myMethod{get\_add$(\mathbf{Y}^*$, $\mathbf{Y})$}{

gt\_skelet $\leftarrow$ skeletonize$(\mathbf{Y}^*)$

false\_negatives $\leftarrow$ gt\_skelet $\land$ $\neg$  $\mathbf{Y}$

\Return false\_negatives
}

\BlankLine
\myMethod{get\_erase$(\mathbf{Y}^*$, $\mathbf{Y}$, $\mu$, $\sigma)$}{

\Comment{dilate gt for more lenient detection}
expanded\_gt $\leftarrow$ dilate$($skeletonize$(\mathbf{Y}^*)$, round$(\mu$ + $2\sigma))$
 
pred\_skelet $\leftarrow$ skeletonize$(\mathbf{Y})$

false\_positives $\leftarrow$ $\neg$expanded\_gt $\land$ pred\_skelet

\Return false\_positives
}
\caption{False Positive/Negative Detection for $\mathbf{\Delta}^-$/$\mathbf{\Delta}^+$}
\label{alg:methods}
\end{algorithm}

Next, for the remaining steps, we propose \Cref{alg:methods,alg:edges}: For (2), we first employ \Cref{alg:methods} to obtain a binary mask of missing or superfluous skeletonized segments, i.e., false positives or negatives.
Note that in order to avoid the correction of minor superfluous parts, in \texttt{get\_erase}, we dilate the ground truth to a constant of $\mu + 2\sigma$ s.t.\ only false positives which drastically diverge from $\mathbf{Y}^*$ will be detected.


\begin{algorithm}

\SetKwComment{Comment}{// }{}
\SetAlgoLined
\KwData{skeletonized mask $\mathbf{S}$}

\BlankLine

$\text{Let } K_\text{edge} = \begin{bmatrix} 
1 & 1 & 1 \\
1 & 10 & 1 \\
1 & 1 & 1
\end{bmatrix}\text{ and } K_\text{label}=\begin{bmatrix} 
1 & 1 & 1 \\
1 & 1 & 1 \\
1 & 1 & 1
\end{bmatrix}.$

\BlankLine

\SetKwProg{myMethod}{def}{:}{}

\myMethod{get\_edges$(\mathbf{S})$}{
conv\_skelet $\leftarrow$ convolve$(\mathbf{S}$, kernel=$K_\text{edge})$

edges $\leftarrow$ conv\_skelet == 12

\BlankLine

\Comment{8-connectivity}
edge\_list $\leftarrow$ label\_connectivity(edges, kernel=$K_\text{label}$)

edge\_list $\leftarrow$ sort(edge\_list, ord='desc')

\Return edge\_list
}

\caption{Obtain Edge Segments, Sorted by Length}
\label{alg:edges}
\end{algorithm}

After obtaining skeletonized binary masks for false positives and negatives, for step (3), we obtain, for both, the longest line segment utilizing \Cref{alg:edges}.
Within \Cref{alg:edges}, we leverage a key property of skeletonizing: In a pixel-based, skeletonized representation (i.e., one where lines have been reduced to their medial axis, which is 1-pixel wide), a single continuous line, will have exactly two neighbors in its 8-neighborhood, except for endpoints and junctions.

Let $\mathbf{\Delta}^+ \in \{0,+1\}^{H,W}$ and $\mathbf{\Delta}^-\in \{0,-1\}^{H,W}$ denote the missing/superfluous line segment that will be added/erased.
Since these operations will be performed interactively, we summarize with $\mathbf{\Delta}$ multiple interactions and thus contains values $\{-1,0,+1\}$.
Finally, we combine previous interactions~$\mathbf{\Delta}$ with~$\mathbf{\Delta}^+$ or~$\mathbf{\Delta}^-$ by leaving previously set values of $\pm1$ fixed, only updating 0-valued values.
For simplicity, we introduce $\mathbf{Y}^\mathbf{\Delta}$, a quantity which denotes the union between the initial prediction $\mathbf{Y}$ and the human interactions $\mathbf{\Delta}$, i.e.:

\begin{equation}
\mathbf{Y}_{i,j}^\mathbf{\Delta}= \begin{cases} 
1 & \text{if } \mathbf{\Delta}_{i,j} == +1 \\
0 & \text{if } \mathbf{\Delta}_{i,j} == -1 \\
\mathbf{Y}_{i,j} & \text{otherwise.}
\end{cases}
\end{equation}

\subsubsection{Interaction}
After introducing the three necessary steps for the adding/erasing operation, we move on to performing realistic human interactions: We continue from the previously found quantities $\mathbf{\Delta}^+$ or~$\mathbf{\Delta}^-$ for adding missing/erasing superfluous segments and either pick one of the two at random during training or the longer segment for maximum correction during inference.
Within the skeletonized segment, we proceed by randomly sampling a sub-segment of up to eleven pixels (a parameter that we did not vary) and dilating it, based on the statistics of~$\mathcal{G}$, with one of the following: (a) a width sampled from the distribution, (b) the mean $\mu$, or (c) a width of $\mu - 2\sigma$.

In \Cref{sec:evaluation}, options~(a) and~(b) will be evaluated w.r.t.\ validation performance, and option~(c) will be used for the final evaluation on whole mirrors s.t.\ human interactions are with high probability aligned with~$\mathbf{Y}^*$, i.e., reduce the risk of strokes being too wide.

\begin{figure}[t]
    \hspace{0.0561445783132\textwidth}
	\centering
		\includegraphics[width=0.8\textwidth]{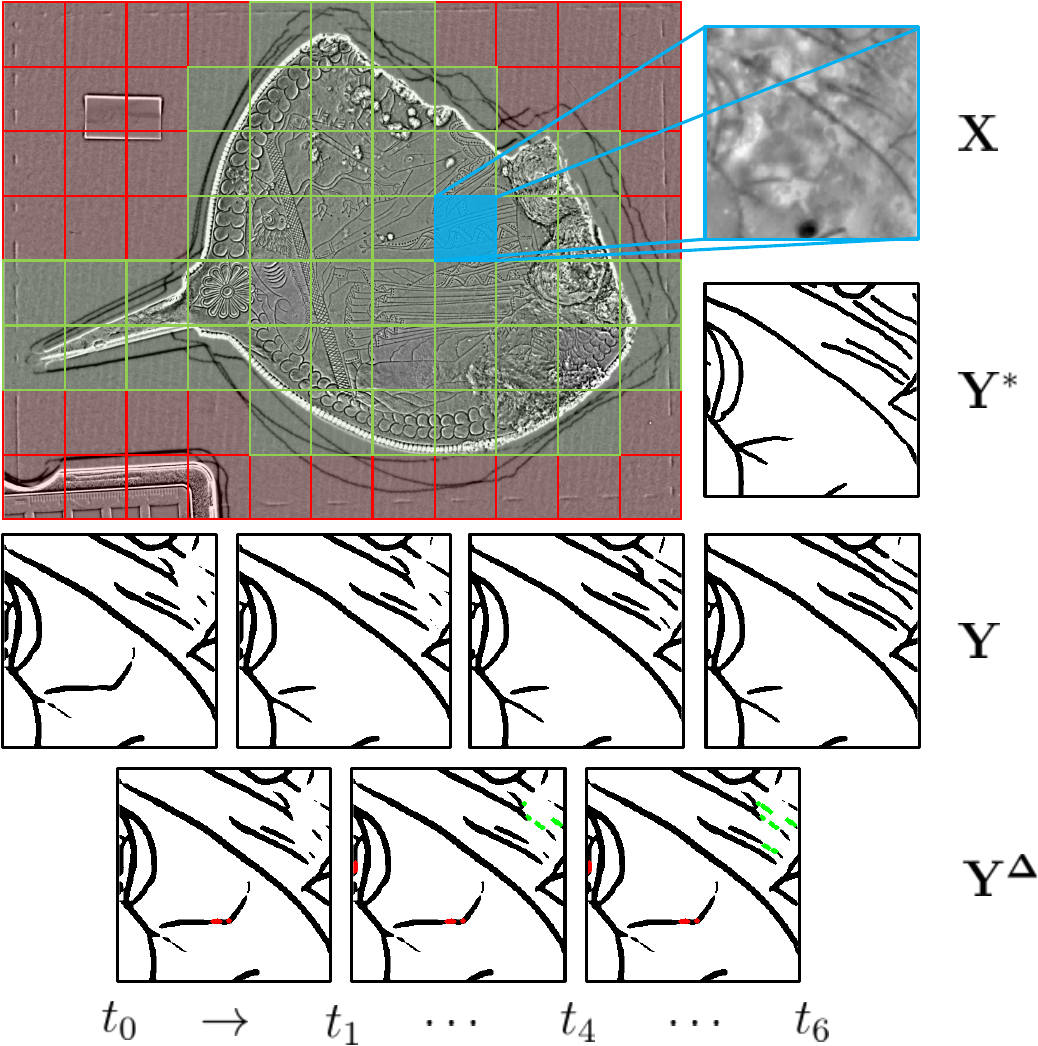}
    \caption{
    An illustration of the overall methodology: In general, segmentation is performed on a per-patch level ($512\times512$, resized to~$256\times256$; red denotes patches that are filtered a priori using SAM~\cite{kirillov_segment_2023}).
    In an interactive paradigm, starting from the initial prediction~$\mathbf{Y}$ at timestep~$t_0$, based on input~$\mathbf{X}$, a human provides hints in the form of~$\mathbf{\Delta}$ (the ``union'' between~$\mathbf{Y}$ and~$\mathbf{\Delta}$ is denoted with~$\mathbf{Y}^\mathbf{\Delta}$), on which a separately trained network $f_{iter}$ is conditioned on to produce a refined mask at timestep $t_1$.
}
    \label{fig:pipeline}
\end{figure}

Finally, using a separate network $f_{iter}$, trained to refine~$\mathbf{Y}$ conditioned on~$\mathbf{\Delta}$ and~$\mathbf{X}$, we obtain a refined prediction~$\mathbf{Y}'$.
With this, we motivate the interactivity of our method: Starting over, i.e., $\mathbf{Y} \leftarrow \mathbf{Y}'$, we again find the segment that requires the most correction and update $\mathbf{\Delta}$ with newly found $\mathbf{\Delta}^+$/$\mathbf{\Delta}^-$. 
A general overview of the interactivity is provided by \Cref{fig:pipeline}, illustrating inference on a per-patch level, the initial prediction $\mathbf{Y}$ and its refinement over time, based on~$\mathbf{\Delta}$.

\subsection{Architecture}
With regards to our architecture, we employ a UNet~\cite{ronneberger_u-net_2015} with an EfficientNet-B6~\cite{tan_efficientnet_2020} following the proposal by Sterzinger et al.~\cite{sterzinger_drawing_2024} but expand upon the input to condition the network on the (simulated) human input $\mathbf{\Delta}$.
For clarification, the input is now comprised of a $3\times H \times W$ tensor, including the depth map $\mathbf{X}$, the human input $\mathbf{\Delta}$, as well as the initial prediction $\mathbf{Y}$ with all three quantities concatenated. 
Given that our data resources are limited, we train on a per-patch-level employing augmentations among which are rotations, flips, and shifts, optimizing the Dice loss.
For the initial prediction $\mathbf{Y}$, we employ the exact same methodology as proposed by Sterzinger et al.~\cite{sterzinger_drawing_2024}. 

\section{Evaluation}
\label{sec:evaluation}
In this section, we evaluate our design choices: During this process, we report the Intersection-over-Union~(IoU) as well as the pseudo-F-Measure~(pFM), a metric commonly used for evaluating the binarization quality of handwritten documents.
It is thus well-suited for our binarization task, i.e., a task where shifting the mask by a single pixel will have a significant impact on per-pixel metrics.
Compared to the standard F-Measure, the pFM relies on the pseudo-Recall~(p-Recall) which is calculated based on the skeleton of $\mathbf{Y}^*$~\cite{Pratikakis2012}:
\begin{equation}\label{eq:pfm}
\text{pFM}(\mathbf{Y}', \mathbf{Y}^*) = \frac{2 \times \text{p-Recall}(\mathbf{Y}', \mathbf{Y}^*) \times \text{Precision}(\mathbf{Y}', \mathbf{Y}^*)}{\text{p-Recall}(\mathbf{Y}', \mathbf{Y}^*) + \text{Precision}(\mathbf{Y}', \mathbf{Y}^*)}
\end{equation}

Given that we work within an interactive paradigm, we are required to also provide a metric that excludes the human input~$\mathbf{\Delta}$ from the evaluation and hence report the relative pFM improvement over~$\mathbf{Y}^\mathbf{\Delta}$, i.e.:

\begin{equation}\label{eq:relpfm}
\text{pFM}_\mathbf{\Delta}(\mathbf{Y}',\mathbf{Y}^\mathbf{\Delta},\mathbf{Y}^*) = \frac{\text{pFM}(\mathbf{Y}',\mathbf{Y}^*)-\text{pFM}(\mathbf{Y}^\mathbf{\Delta},\mathbf{Y}^*)}{\text{pFM}(\mathbf{Y}^\mathbf{\Delta},\mathbf{Y}^*)}
\end{equation}

In addition, based on the fact that during training we introduce randomness,~i.e., by chance, missing parts can be added~($\mathbf{\Delta}^+$) or superfluous ones erased~($\mathbf{\Delta}^-$), and that sub-segments are sampled and dilated at random, we evaluate on the test/validation set five times and report the average. 

\subsection{Training}
Our model $f_{iter}$ is trained on an NVIDIA RTX~A5000 until convergence,~i.e., no improvement~$\geq 1e-3$ w.r.t.\ the $\text{pFM}_\mathbf{\Delta}$ (see Equation~\ref{eq:relpfm}) for ten consecutive epochs, using a batch size of 32 and a learning rate of $3e-4$.
As a loss function, we employ a generalized Dice overlap (Dice loss) that is well suited for highly unbalanced segmentation masks~\cite{sudre_generalised_2017} and optimize it using Adam~\cite{kingma_adam_2017}.
Additionally, we incorporate a learning rate scheduler that also monitors the~$\text{pFM}_\mathbf{\Delta}$ on our validation set: If there is no improvement for three consecutive training epochs, the learning rate is halved. 

\subsection{Ablation Study}
In the following, we present our ablation study, focusing on input options, different stroke widths~(widths kept fixed and sampled randomly), as well as the necessity of our two operations~(add and erase).

\begin{table}
\centering
\caption{
Evaluating input options and their effect on the per-patch predictive performance (fixed stroke width, one interaction): Iterating over $\mathbf{Y}$ again does not cause improvement whereas providing $\mathbf{\Delta}$ yields ca.\ $+6\%$ over $\mathbf{Y}^\mathbf{\Delta}$.
Note that, although part of the input, we hide~$\mathbf{X}$ for clarity.
}
\begin{tabular*}{\linewidth}{lc@{\extracolsep{\fill}}llc}
\toprule
    \multicolumn{2}{l}{\textbf{Input Modality}} & \multicolumn{1}{l}{\textbf{IoU}} & \multicolumn{1}{l}{\textbf{pFM}} & \multicolumn{1}{c}{\textbf{$\text{pFM}_\mathbf{\Delta}$}}\\
\midrule
Init. Prediction~\cite{sterzinger_drawing_2024} &-&32.86&49.28&$-$\\
\midrule
Prediction &$\mathbf{Y}$&32.72&49.27&$-$\\
Interaction  &$\mathbf{\Delta}$&$35.83\pm.1$&$53.60\pm.2$&$+5.8\pm.23\%$\\
\midrule
Both               &$\mathbf{Y},\mathbf{\Delta}$&$36.04\pm.2$&$53.44\pm.3$&$+5.5\pm.55\%$\\
\bottomrule
\end{tabular*}
\label{tab:input}
\end{table}

\paragraph{Input Options:} Starting with the evaluation of different input options and their impact on the predictive patch-wise performance of $f_{iter}$ (fixed stroke width, one interaction; results are denoted in \Cref{tab:input}): As expected, simply iterating over the initial prediction $\mathbf{Y}$ (stemming from network $f_{init}$) results in no improvement, rendering the human an essential part of the refinement process.
Moreover, by means of human guidance, i.e., providing $\mathbf{\Delta}$, the network can effectively leverage additional information on missing or superfluous parts, resulting in an increase of around +6\% over $\mathbf{Y}^\mathbf{\Delta}$.
Finally, when utilizing both~$\mathbf{Y}$ and~$\mathbf{\Delta}$, we attain a comparable improvement over $\mathbf{Y}^\mathbf{\Delta}$, with the difference deemed not statistically significant at a confidence level of 95\%.
However, the latter results in faster convergence, the reason for which we proceed with this option.

\paragraph{Stroke Widths:} Next, we consider different options for the stroke width during the simulation of human interaction, namely:~(a) keeping the stroke width constant at~$\mu$ and~(b) sampling it from~$\mathcal{G}$.
Our evaluation reveals that sampling does not significantly improve performance, with results showing $+5.5\pm.55\%$ improvement for fixed width versus $+4.9\pm.35\%$ for sampled one.

\begin{table}
\centering
\caption{
Evaluating the impact of adding and erasing when refining mirror ANSA-1700: Employing both operations will result in the highest $\text{pFM}_\mathbf{\Delta}$ of ca.~+12\%, where adding has a greater impact (ca.~+8\%) than erasing (ca.~+2\%); note that results are reported at the maximum $\text{pFM}_\mathbf{\Delta}$.
}

\begin{tabular*}{\linewidth}{l@{\extracolsep{\fill}}cccr}
\toprule
    \multicolumn{2}{l}{\textbf{Interaction}} & \multicolumn{1}{c}{\textbf{IoU}} & \multicolumn{1}{c}{\textbf{pFM}} & \multicolumn{1}{c}{\textbf{$\text{pFM}_\mathbf{\Delta}$}}\\
\midrule
Only Erasing&$\mathbf{\Delta}^-$ &$38.25\pm.06$&$58.92\pm.07$&$+1.9\pm.12\%$\\
Only Adding&$\mathbf{\Delta}^+$ &$55.19\pm.13$&$73.55\pm.11$&$+8.4\pm.16\%$\\
\midrule
Both&$\mathbf{\Delta}^{\phantom{+}}$&$58.41\pm.28$&$76.56\pm.16$&$+12.3\pm.17\%$\\
\bottomrule
\end{tabular*}
\label{tab:human_interaction}
\end{table}

\begin{figure}
        \centering
		\includegraphics[width=0.85\textwidth]{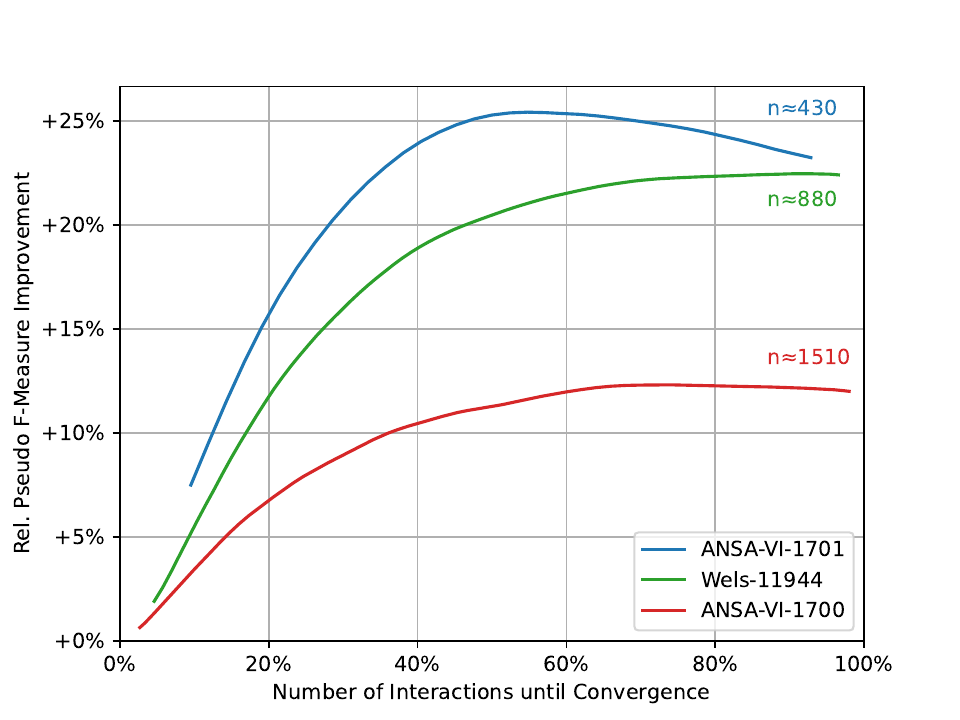}
        \caption{An illustration of $\text{pFM}_\mathbf{\Delta}$, i.e., the relative pFM improvement of our method over pure manual refinement; $n$ denotes the number of human interactions: With the relative improvement peaking at values between ca.~+12\% and +26\%, our human-in-the-loop approach immediately overtakes manual labeling, leading to drastically better annotations earlier.}
        \label{fig:rel_impro}
\end{figure}

\paragraph{Operations:}
\label{sec:steps}
In order to illustrate the necessity of our two operations, namely adding \( \mathbf{\Delta}^+ \) and erasing \( \mathbf{\Delta}^- \), we perform multiple interactions until convergence and report results at the maximum attained \( \text{pFM}_\mathbf{\Delta} \).
For this, we inspect an entire mirror, ANSA-1700: Employing both operations jointly yields the highest \( \text{pFM}_\mathbf{\Delta} \) of approximately +12\% (redline in ~\Cref{fig:rel_impro}).
Notably, \textit{add} has a more significant impact (ca.~+8\%) compared to \textit{erase} (ca.~+2\%).
However, this is very dependent on the initial prediction, thus only demonstrating that one operation supplements the other.

In summary, compared to the initial prediction $\mathbf{Y}$, stemming from $f_{init}$, providing human guidance via $\mathbf{\Delta}$ will yield improvements exceeding $\mathbf{Y}^\mathbf{\Delta}$, utilizing both operations is beneficial, and augmenting stroke widths by random sampling performs worse than leaving it constant.

\section{Results}
\begin{figure}
        \centering
		\includegraphics[width=0.85\textwidth]{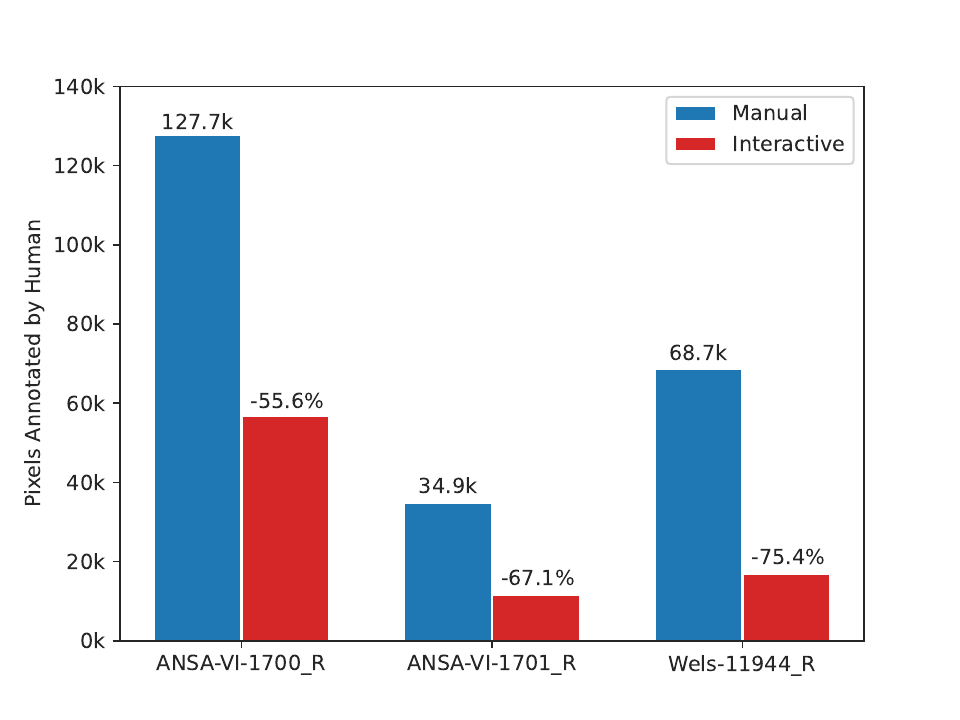}
        \caption{An illustration of reduced workload: At convergence, our interactive approach requires drastically fewer annotated pixels to reach equal performance in pFM, resulting in a reduction of annotation effort ranging from 56\% to 75\%.}
        \label{fig:workload_reduction}
\end{figure}

After verifying the effectiveness of our methodology, we pick the three mirrors from our validation/test set, namely ANSA-1700, ANSA-1701, and Wels-11944, and evaluate our human-in-the-loop approach on whole mirrors, performing multiple interactions (limited to 3,000; typically requiring much less). 
Again, due to the introduced randomness, we repeat this process ten times and report the average result, skipping the variation as it is negligible.
As described in \Cref{sec:steps}, for this, we start greedily by selecting the patch with the lowest pFM, simulate adding missing/erasing superfluous parts, selecting the operation which yields a larger improvement, refine the prediction based on the additional human input, and proceed from there until \textit{convergence}, i.e., when neither adding nor erasing by itself increases the metric.
We report the results of this in two figures: \Cref{fig:rel_impro}, which illustrates the relative pFM improvement over-performed interactions, as well as \Cref{fig:workload_reduction}, which depicts potential reduction in annotation workload when employing our proposed interactive refinement paradigm.

Inspecting \Cref{fig:rel_impro}, we observe for all three mirrors significant relative improvements over the purely manual annotation baseline $\mathbf{Y}^\mathbf{\Delta}$ when employing our method, with maximum improvements ranging from +12\% to +26\% depending on the mirror under consideration.
Based on these results, we conclude that our human-in-the-loop approach quickly overtakes manual labeling, leading to drastically better annotations at an earlier stage.
Interestingly, towards the end, relative improvement starts to decrease slightly before convergence (most notably for ANSA-1701), showcasing that at a point, the network will undo previously correctly annotated parts.

In \Cref{fig:workload_reduction}, we directly contrast pure manual refinement against our interactive approach.
For this, we determine the maximum attained pFM, calculated using~$\mathbf{Y}^\mathbf{\Delta}$, which corresponds to the final simulated human interaction.
We then compare the amount of required human input to the human input necessary to reach equal or higher performance using our proposed method.
By doing this, we are able to report a notion of workload reduction: Depending on the mirror under inspection, annotation requirements will experience a reduction ranging from around -56\% to -75\%, positioning our model well to be employed for simplifying the task of correcting erroneous segmentation masks.

\section{Limitations and Future Work}
While our proposed method shows promising results, it is important to acknowledge its limitations: At the moment human guidance aids refinement only locally,~i.e., modifications happen just in the vicinity of the provided annotation.
Moving forward, one could focus on further refining our methodology by exploring additional techniques to enhance efficiency.
For instance, it would be meaningful to investigate the integration of quickly trainable learning algorithms, such as Gaussian processes which can immediately be adapted to newly provided annotation and thus allow for global adjustments, potentially further reducing the amount of human input required.
Additionally, leveraging Gaussian processes is accompanied by the option of active learning strategies, which could allow the identification and annotation of patches where the model is most uncertain with the chance of expediting refinement further.

\section{Conclusion}
In summary, our research addresses the labor-intensive process of manually tracing intricate figurative illustrations found, for instance, on ancient Etruscan mirrors.
In an attempt to automate this process, previous work has proposed the use of photometric-stereo scanning in conjunction with deep neural networks.
By doing so, quantitative performance comparable to expert annotators has been achieved; however, in some instances, they still lack precision, necessitating correction through human labor.
In response to the remaining resource intensity, we proposed a human-in-the-loop approach that streamlines the annotation process by training a deep neural network to interactively refine existing annotations based on human guidance.
For this, we first developed a methodology to mimic human annotation behavior: We began by analyzing annotation statistics to capture stroke widths accurately and proceeded by introducing algorithms to select erroneous patches, identify false positives and negatives, as well as correct them by erasing superfluous or adding missing parts.
Next, we verified our design choices by conducting an ablation study; its results showed that providing human guidance will yield improvements exceeding pure manual annotation, utilizing both operations is beneficial, and augmenting stroke widths by random sampling performs worse than leaving it constant.
Finally, we evaluated our method by considering mirrors from our test and validation set.
Here, we achieved equal quality annotations with up to 75\% less manual input required.
Moreover, the relative improvement over pure manual labeling reached peak values of up to 26\%, highlighting the efficacy of our approach in reaching drastically better results earlier.

%
%
%
\bibliographystyle{splncs04}
\bibliography{bibliography}

\end{document}